\documentclass[runningheads]{llncs}

\usepackage{algorithm}
\usepackage{algpseudocode}
\usepackage{amssymb}
\usepackage{amsmath}
\usepackage{graphicx}
\usepackage{subfig}
\usepackage{pgfplots}
\usepackage{soul}
\usepackage{cite}
\pgfplotsset{compat=1.18}

\begin{document}

\title{Accelerating Machine Learning Primitives on Commodity Hardware}
\iftrue
\author{Roman Snytsar\orcidID{0000-0001-9042-8841}}

\authorrunning{R. Snytsar}

\institute{AI \& Research, Microsoft, Redmond WA 98052, USA \\
\email{Roman.Snytsar@microsoft.com}}
\fi

\maketitle

\begin{abstract}
{Sliding Window Sum algorithms have been successfully used for training and inference of Deep Neural Networks. We have shown before how both pooling and convolution 1-D primitives could be expressed as sliding sums and evaluated by the compute kernels with a shared structure. \\
In this paper, we present an extensive study of the Sliding Window  convolution technique as a more efficient alternative to the commonly used General Matrix Multiplication (GEMM) based convolution in Deep Neural Networks (DNNs). The Sliding Window technique addresses the memory bloating problem and demonstrates a significant speedup in 2-D convolution. We explore the performance of this technique on a range of implementations, including custom kernels for specific filter sizes.\\
 Our results suggest that the Sliding Window computation kernels can outperform GEMM-based convolution on a CPU and even on dedicated hardware accelerators. This could promote a wider adoption of AI on low-power and low-memory devices without the need for specialized hardware. We also discuss the compatibility of model compression methods and optimized network architectures with the Sliding Window technique, encouraging further research in these areas.
}

\end{abstract}
\vspace{0.35cm}

\section{Introduction}

In recent years, there has been significant progress in machine learning (ML) research, with breakthroughs in deep learning, natural language processing, and computer vision. A Deep Neural Network (DNN) is one of the most significant tools of a ML scholar \cite{li2021survey}. DNNs are constructed from multiple layers that transform the data sequentially via operations such as pooling, convolution, and activation. 
In most successful DNNs, the greater portion of computational resources is consumed by performing convolution.\\

A popular implementation of convolutional layers is expanding the input into a column matrix form (im2col) and then calling a highly tuned General Matrix Multiplication (GEMM) procedure from the existing linear algebra library such as BLIS \cite{van2015blis} or MKL \cite{wang2014intel}. Since the hardware optimized GEMM implementations exist for every standard CPU, graphics processing unit (GPU), or digital signal processor (DSP), the im2col approach has been highly successful in DNN frameworks such as Caffe \cite{jia2014caffe}, Torch \cite{collobert2002torch} and ONNX \cite{onnx2023}.\\

However, these advances have primarily benefited large corporations, and research institutions with access to massive computational resources. 
 The democratization of AI on low power and edge devices aims to bring the benefits of AI to a wider audience, including small businesses, individual users, and the billions of Internet of Things (IoT) devices. Edge devices, such as smartphones, wearables, and IoT sensors, are often resource-constrained, with limited processing power, memory, and battery life.\\
 
One major challenge in deploying AI on edge devices is the size of deep learning models, which can be hundreds of megabytes or even gigabytes. The im2col conversion further increases the memory footprint of the input matrix and reduces data locality. For a convolution with a filter size k, the column matrix is k times larger than the original input tensor. A lot of research effort has been put into applying the GEMM routines to the smaller intermediate data structures \cite{anderson2017low} \cite{vasudevan2017parallel} or even to the original input data \cite{wang2021optimization}.\\

To reduce the memory requirements on edge devices and improve performance, researchers have been exploring various techniques, including model compression, network optimization, and hardware acceleration.

\subsection{Model Compression}
Model compression techniques aim to reduce the size of these models while maintaining their accuracy. Various methods have been proposed, including quantization, weight pruning, and knowledge distillation.\\

Quantization is a technique that reduces the numerical precision of model weights and activations to the low precision floating point or even integer representation \cite{gholami2021survey}. The quantization reduces the memory footprint and latency by an order of magnitude with only a minor degradation in accuracy for many tasks. This is particularly important for edge devices, where memory and computational resources are limited. There are also several types of extreme quantization, such as binary and ternary quantization, which use only two or three discrete values for weights and activations \cite{courbariaux2015binaryconnect} \cite{hubara2016binarized}, and mixed-precision quantization, which uses different levels of precision for different parts of the network \cite{choi2018pact} \cite{zhou2016dorefa}. It is important to note however, that quantization is not entangled with GEMM and could be equally successful when applied to the original convolution problem.\\

Pruning is a method that removes redundant or less important parameters from the model, resulting in a sparse model with less parameters \cite{han2015learning}). This can lead to significant reductions in memory and computational requirements, making the model more suitable for deployment on edge devices. Pruning techniques can be broadly classified into unstructured pruning, which removes individual weights, and structured pruning, removing entire neurons, filters, or channels \cite{wen2016learning}. Unstructured pruning can achieve high levels of sparsity but may not lead to actual speedup on hardware, whereas structured pruning can result in a more efficient hardware implementations. Dynamic pruning methods have also been developed, which adjust the sparsity of the model during training or inference, further improving efficiency \cite{guo2016dynamic} \cite{lin2017deep}.\\

Knowledge distillation is a technique that trains a smaller model (student) to mimic the behavior of a larger, pre-trained model (teacher), effectively transferring knowledge from the teacher to the student \cite{hinton2015distilling}. This can result in a compact model that achieves similar accuracy to the larger model, making it more suitable for deployment on edge devices. Knowledge distillation typically involves training the student model using a combination of the original task loss and a distillation loss, which measures the discrepancy between the teacher and student model outputs. Various distillation losses have been proposed, such as the Kullback-Leibler divergence between the softened output probabilities, or the mean squared error between feature maps or logits.
\subsection{Optimized Network Architectures}
Further reducing the number of parameters and computations, optimized network architectures help bring the benefits of AI to a wider range of applications  and devices, while addressing challenges such as latency, privacy, and energy efficiency.\\

MobileNets are a family of efficient convolutional neural networks designed for mobile and embedded vision applications. They are based on depthwise separable convolutions, which significantly reduce the number of parameters and computations compared to standard convolutions \cite{howard2017mobilenets}. MobileNetV2 \cite{sandler2018mobilenetv2} incorporates inverted residual structures and linear bottlenecks for improved efficiency.\\

EfficientNets are a series of neural networks derived using a combination of neural architecture search and compound scaling. They achieve state-of-the-art accuracy with significantly fewer parameters compared to other models of similar accuracy. EfficientNets have been successfully used for various computer vision tasks \cite{tan2019efficientnet}.\\

ShuffleNet is an efficient neural network architecture that employs pointwise group convolutions and channel shuffling to reduce computation while maintaining accuracy. Pointwise convolution is equivalent to matrix multiplication and thus extremely efficient as has been proven by the ShuffleNet application to the assortment of image classification tasks \cite{zhang2018shufflenet}.\\

SqueezeNet is a compact neural network architecture that achieves AlexNet-level accuracy on the ImageNet dataset with 50x fewer parameters \cite{iandola2016squeezenet}. It uses "fire modules" consisting of a squeeze convolution layer followed by an expansion layer to reduce the number of parameters while maintaining accuracy.
\subsection{Hardware Accelerators}
Custom accelerators  are specialized hardware units designed specifically for the efficient execution of neural networks. These accelerators often feature optimizations such as low-precision arithmetic, dataflow-based execution, and on-chip memory hierarchies to minimize energy consumption and latency. Examples of custom accelerators for edge devices that borrow their designs from the high-power applications include Google's Edge TPU \cite{jouppi2017datacenter} and NVIDIA's Deep Learning Accelerator (DLA) \cite{nvidia2018}. These are in fact the GEMM accelerators as the popularity of the im2col tactics significantly influences the design of the custom hardware.\\

Graphics Processing Units (GPUs) have evolved from specialized hardware for rendering graphics to highly parallel and programmable architectures suitable for a wide range of high-performance computing. They are widely used for accelerating neural network. In recent years, GPU manufacturers have introduced low-power GPUs specifically designed for edge devices, such as NVIDIA's Jetson platform \cite{nvidia2021} and the ARM Mali-G series \cite{arm2021}. Following the high-power GPU designs, the trend is reversing from general purpose compute to more specialized “tensor cores” for GEMM operations.\\

Field-Programmable Gate Arrays (FPGAs) are reconfigurable hardware platforms that can be programmed to implement custom logic circuits, making them suitable for accelerating neural networks on edge devices. FPGAs offer flexibility and can be reprogrammed to adapt to different models or tasks. Examples of FPGA platforms for edge devices include Intel's Movidius Myriad X \cite{intel2021} and Xilinx Zynq UltraScale+ \cite{xilinx2021}.

\section{Experiments}
Earlier we proposed a new algorithm for performing convolution. The Sliding Window technique \cite{snytsar2020parallel} replaces GEMM with a novel computation kernel that operates on the unmodified input and eradicates the memory bloating problem. The speedup of 1-D convolution we have observed when compared to the baseline \emph{MlasConv} procedure was roughly proportional to the logarithm on the filter width \cite{snytsar2023sliding}.\\

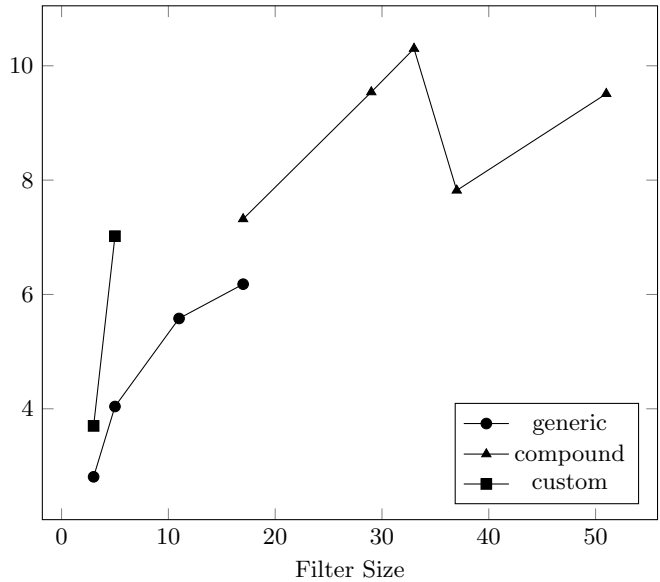
\begin{figure}[!tpb]
	\centering
	\begin{tikzpicture}
		\begin{axis}[
            width=.8\textwidth,
			xlabel={Filter Size},
   legend pos=south east]
			\addplot[mark=*] coordinates {
(3,	2.81)
(5,	4.04)
(11,	5.58)
(17,	6.18)
			};
   \addlegendentry{generic}
   \addplot[mark=triangle*] coordinates {

(17,	7.32	)
(29,		9.54	)
(33,		10.3	)
(37,		7.82	)
(51,		9.51	)
			};
      \addlegendentry{compound}
   \addplot[mark=square*] coordinates {
(3,	3.7)
(5,7.02)

			};
      \addlegendentry{custom}
		\end{axis}
	\end{tikzpicture}
	\caption{Speedup of the 2-D Convolution.}\label{speedup1d}
\end{figure}
In this paper we present the extension of the Sliding Window algorithms to the more practical 2-D cases. There are three  different implementations on the 2-D sliding convolution. The kernel sizes up to 17 are handled by the straightforward version of the Vector Slide algorithm. Kernels of larger width  do not fit into the hardware vector and require a special version that operates on multiple hardware vectors treating them as a single long compound vector. Both generic versions perform redundant shuffles, so  for filter widths 3 and 5  we implemented custom kernels with optimal number of operations.\\

We have run experiments on an Azure node with 16 cores of Intel(R) Xeon(R) Platinum 8272CL CPU and 32 GB of RAM. The speedup is measured compared to the ONNX \emph{MlasConv} calls. All tests have been run in a single-core configuration to exclude the effects of the task scheduling delays.\\

The 2-D Sliding Window convolution exhibits the same roughly logarithmic speedup in correlation to the filter size. The zigzag pattern at the larger filter sizes is related to the alignment of the compound vector to the hardware vector length.\\

Custom implementations are indeed faster than their generic counterparts. Generating custom kernels at run time might improve the performance for every filter size.\\

An interesting observation happens at filter size 17 as it could be handled by either hardware-specific or compound implementation. The compound variation is significantly faster. It is worth studying this phenomenon closer in hopes of improving the performance of the hardware-specific code and bringing the whole left part of the graph higher.\\

The number of arithmetic operations performed by the sliding convolution is the same as the naïve or GEMM-based algorithms. Our observations hint that the speedup comes from better memory access patterns. \\

We have also measured the arithmetic throughput of different kernels using the Intel Advisor \cite{o2017intel}. As the filter size increases, the throughput of the Sliding Window convolution kernels approaches the hardware limits. It is also interesting to see that the filter size misalignment with the hardware vector length results in similar performance patterns for both Sliding Window and GEMM kernels.\\
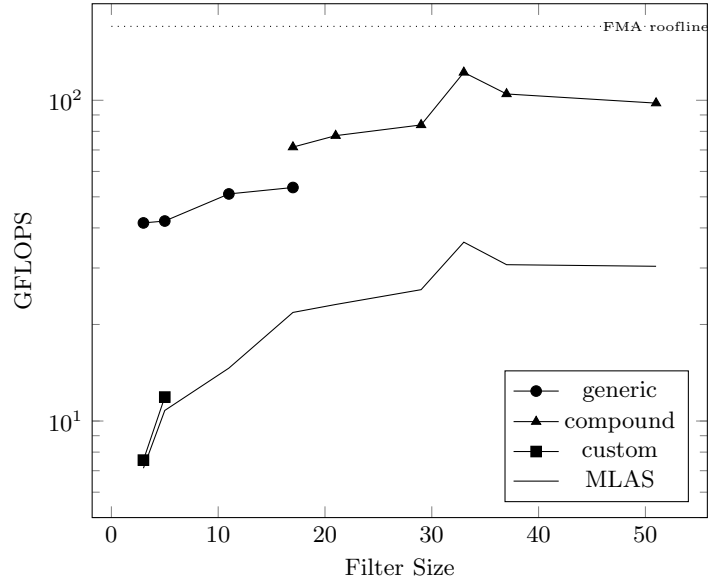
\begin{figure}[!tpb]
	\centering
	\begin{tikzpicture}
		\begin{axis}[
            width=.8\textwidth,
            ymode=log,
			xlabel={Filter Size},
            ymin=5,
            ymax = 200,
			ylabel={GFLOPS},
   legend pos=south east]
			\addplot[mark=*] coordinates {
(3,	41.461)
(5,	42.028)
(11,	51.048)
(17,	53.465)
			};
   \addlegendentry{generic}
			\addplot[mark=triangle*] coordinates {
(17,	71.466)
(21,		77.576)
(29,		83.784)
(33,		122.135)
(37,		104.638)
(51,		97.978)
			};
      \addlegendentry{compound}
			\addplot[mark=square*] coordinates {
(3,		7.554)
(5,		11.877)
			};
      \addlegendentry{custom}
			\addplot[mark=none] coordinates {
(3,	7.128)
(5,	10.789)
(11,	14.62)
(17,	21.806)
(21,	23.086)
(29,	25.675)
(33,	36.11)
(37,	30.722)
(51,	30.37)
			};
      \addlegendentry{MLAS}
      \draw [dotted] (axis cs:0,170) -- (axis cs:51,170) node[font=\tiny]{FMA roofline};
		\end{axis}
	\end{tikzpicture}
	\caption{2-D Convolution throughput.}\label{dilated1d}
\end{figure}
\section{Conclusion}
We have measured the performance and throughput of the Sliding Window convolution kernels.  They are more efficient than the commonly used GEMM kernels on the CPU and could even outperform dedicated hardware accelerators. Wider adoption of the Sliding Window sum algorithm could promote AI usage on the low power and low memory devices avoiding the expense of specialized hardware.\\

All the model compression techniques  described earlier are equally applicable to Sliding Window computation. Pruning and distillation reduce the work required for ML inference. Quantization delivers the same benefits of memory and power savings, and better vector performance.\\

Optimized network architectures tend to use small convolution filters that diminish the advantages of the Sliding Window convolution. 
In the extreme case  of ShuffleNet its pointwise convolutions do not benefit from the new algorithm at all. In general, small filter convolutions are memory bound, equally limiting performance of custom accelerators and CPU solutions. We encourage new research into the network architectures that use fewer layers with larger convolution filters.\\

In many cases the hardware accelerators can be repurposed for Sliding Window algorithms with various degree of success depending on how specialized the hardware is.\\

Pipelined nature of the Sliding Window algorithm ensues a straightforward FPGA implementation. Limited on-chip memory and logic resources can be a constraint for implementing large-scale deep learning networks. Combining Sliding Window techniques with optimized network architectures and model compression results in fast and energy efficient solutions.\\

The algorithms are easily portable to GPU as well. The benefits of streamlined memory access are less pronounced since explicitly controlled on-chip memory hierarchies make GPUs already highly efficient in GEMM computation.\\

Since the accelerators for matrix multiplication are already present in the current generation of hardware and are likely to stay in future devices, they could improve throughput and performance of many computational tasks beyond GEMM. Thus, it is important to re-formulate our algorithms in terms of the small matrix multiplication, completing the circle. Competition between CPU algorithms and hardware accelerators would lead to advances in both directions, and the most spectacular results are expected at the intersection of the two research fields.\\

Sliding Window convolution algorithms exhibit excellent performance using commodity hardware. They deliver the benefits of AI to more low-cost and low-power devices.

\bibliographystyle{splncs04}
\bibliography{ConvolutionCommodityHardware}
\end{document}